# Stabilizing Trajectory Outputs in End-to-End Autonomous Driving via SC-IMM Based Teacher Signals

Siewoo Kim[1], Seung-Hyun Kong[1,*]

[1]The CCS Graduate School of Mobility, Korea Advanced Institute of Science and Technology

**Abstract:** End-to-End autonomous driving models commonly predict future waypoints from sensor inputs and convert them into vehicle control commands through a downstream controller. However, conventional waypoint-based imitation learning mainly minimizes coordinate-level errors, making it difficult to capture scene-dependent path-speed changes and temporal instability across waypoint outputs. In this paper, we propose an offline teacher-signal generation and learning method for trajectory-output stabilization based on a Scene-Conditioned Interacting Multiple Model (SC-IMM) to mitigate this issue. The proposed method converts expert trajectories into path-speed states and performs IMM updates conditioned on scene cues to generate path-speed teacher labels and mode posterior probabilities. The generated signals are added to the original trajectory loss as auxiliary supervision during training, while the inference structure and waypoint controller remain unchanged. In closed-loop evaluation on 100 short routes in CARLA Town12, the proposed method improved the driving score by 28.0% and reduced Collision/km by 62.3% compared with the baseline, while also improving jerk and trajectory-variation metrics. These results demonstrate that offline teacher signals embedding scene-conditioned motion-model cues can guide trajectory-output driving models toward more stable closed-loop behavior.

## I. INTRODUCTION

Autonomous driving systems have traditionally been developed around modular architectures that separate perception, prediction, planning, and control. More recently, end-to-end autonomous driving, which directly generates driving trajectories or control outputs from sensor inputs, has been actively studied [1-5]. Among these approaches, architectures that predict a future driving trajectory and convert it into control commands through a controller have been widely adopted in recent end-to-end autonomous driving models because they represent the driving plan more explicitly than methods that directly output control commands [6,7].

The future trajectory predicted by a model is converted into steering, acceleration, and braking commands through a controller during closed-loop driving. Therefore, temporal changes in trajectory outputs over a continuous driving sequence are directly related to vehicle-control stability. Even when the average position error is small, large changes in the direction, curvature, or speed profile of predicted trajectories at consecutive time steps can produce discontinuous control inputs and unstable vehicle behavior. In a closed-loop environment, the current prediction and control result also affect the vehicle's next state and observation; consequently, open-loop coordinate error alone cannot adequately explain instability during actual driving [8]. Recent work has likewise reported the importance of temporal consistency in the trajectory outputs of end-to-end autonomous driving models [9].

Conventional imitation-learning methods primarily minimize coordinate-level errors between expert and predicted trajectories. Although this approach is effective for imitating trajectories in the given data, it does not sufficiently capture the continuous trajectory changes and control stability required for closed-loop driving. During actual driving, the direction and speed characteristics of a future trajectory change jointly according to road geometry, the target route, surrounding objects, and speed variation, and these changes are difficult to explain using the coordinate error at a single time step. Privileged-information teacher learning [10], global-context modeling through sensor fusion [11], and learning from the driving experience of both the ego vehicle and surrounding vehicles [12] further demonstrate the importance of scene context and learning-signal design. In learning-based driving, the experience and auxiliary learning signals used during training can also affect model performance [13]. Stabilizing the trajectory outputs of an end-to-end autonomous driving model therefore requires the learning process to consider context-dependent trajectory-change characteristics beyond simple coordinate imitation.

Accordingly, this paper proposes a method that generates teacher signals using a scene-conditioned interacting multiple model (SC-IMM) and applies them to trajectory learning in an end-to-end

* Corresponding Author

Siewoo Kim: Graduate Student, The CCS Graduate School of Mobility, Korea Advanced Institute of Science and Technology (siewoo.kim@kaist.ac.kr ORCID 0009-0008-7026-0453)

Seung-Hyun Kong: Professor, Korea Advanced Institute of Science and Technology (skong@kaist.ac.kr, ORCID 0000-0002-4753-1998)

This work was supported by the Korea Evaluation Institute of Industrial Technology (KEIT) grant funded by the Korean Government (Ministry of Trade, Industry and Energy) (RS-2025-25451359).

autonomous driving model. The proposed method converts expert driving trajectories into path-speed states that jointly represent path and speed characteristics, and performs mode-specific Kalman filter updates within an interacting multiple model (IMM) [14,15]. Rather than relying only on fixed transition probabilities, SC-IMM adjusts mode-transition probabilities and process noise according to driving-context cues such as route commands, curvature, hazards, and speed reduction. This process generates path teacher labels, speed teacher labels, and mode posterior probabilities, which are collectively referred to as SC-IMM-based teacher signals in this paper.

The generated SC-IMM-based teacher signals are used only as auxiliary learning signals during training. The path and speed teacher labels serve as learning targets for trajectory-output training, while the mode posterior probabilities provide soft supervision for auxiliary mode learning. During inference, the original waypoint-output architecture is retained without additional online filtering or controller modifications.

The main contributions of this study are as follows. First, we propose an SC-IMM-based offline teacher-signal generation method that produces path teacher labels, speed teacher labels, and mode posterior probabilities from expert driving trajectories and driving-context cues. Second, we incorporate path-speed changes and driving-mode information into conventional trajectory learning by using the generated teacher signals as auxiliary supervision. Third, through comparisons with a baseline trajectory-learning model and a fixed-transition-probability IMM-based model, we validate the improvements in closed-loop driving performance and trajectory-output stability produced by the proposed method in the CARLA simulator.

The remainder of this paper is organized as follows. Section II reviews related work on end-to-end autonomous driving, closed-loop evaluation and driving stability, and IMM-based multi-mode state estimation. Section III describes the proposed SC-IMM-based teacher-signal generation procedure and learning architecture. Section IV analyzes the characteristics of the teacher signals and closed-loop driving performance, and Section V concludes the paper.

## II. BACKGROUND AND RELATED WORK

### 1. End-to-End Autonomous Driving

End-to-end autonomous driving is a learning-based approach that directly generates a driving trajectory or control-related outputs from sensor inputs. Chen et al. [4] comprehensively reviewed research trends in end-to-end autonomous driving, while Jo and Park [5] combined a reinforcement-learning module with a trajectory-guided control-prediction model to mitigate limitations in scenario adaptation. Wu et al. [6] proposed an architecture that jointly learns future-trajectory and control prediction, and Renz et al. [7] presented a model that improves closed-loop driving performance using only camera inputs. Chen et al. [10], Prakash et al. [11], and Chen and Krähenbühl [12] respectively demonstrated the importance of scene context and learning-signal design through privileged-information teacher learning, camera-LiDAR sensor fusion, and the use of driving experience from the ego vehicle and surrounding vehicles. Codevilla et al. [16] proposed a route-command-conditioned imitation-learning architecture and later analyzed data-bias and generalization limitations in behavior-cloning-based autonomous driving [17]. These studies primarily focused on improving closed-loop driving performance through better input representations, model architectures, and output formats. In contrast, this study focuses on the temporal stability of the future trajectories produced by an end-to-end autonomous driving model and proposes using SC-IMM-based teacher signals to assist trajectory learning.

### 2. Closed-Loop Evaluation and Driving Stability

The performance of an autonomous driving model cannot be adequately assessed using open-loop coordinate error alone. Open-loop evaluation measures the error between predicted and ground-truth trajectories in recorded data, but it does not reflect how the predictions affect the vehicle's actual motion and subsequent observations. Codevilla et al. [18] noted that offline evaluation results for vision-based driving models may not sufficiently explain actual driving performance. To address this limitation, simulation-based closed-loop environments such as CARLA are widely used to validate autonomous driving algorithms [19]. Jia et al. [8] proposed the Bench2Drive benchmark to evaluate a broad range of closed-loop driving capabilities in end-to-end autonomous driving systems.

In trajectory-output-based end-to-end autonomous driving, the predicted future trajectory is converted into vehicle-control commands through a controller. Temporal inconsistency in trajectory outputs can therefore destabilize the control input and is directly related to closed-loop driving stability. Song et al. [9] demonstrated the importance of considering temporal consistency in trajectory outputs for end-to-end autonomous driving. From this perspective, the present study analyzes closed-loop driving metrics such as driving score, route completion, and collision rate together with trajectory and control-stability metrics including steering jerk, longitudinal jerk, curvature variation, and speed variation.

### 3. IMM-Based Multi-Mode State Estimation

The Kalman filter [15] is a representative probabilistic filtering method that recursively estimates the state of a dynamic system from noisy observations. Blom and Bar-Shalom [14] proposed the interacting multiple model (IMM), which maintains multiple mode hypotheses in parallel and fuses their state estimates according to their posterior probabilities. The IMM mixes the mode-specific states and probabilities from the previous time step according to transition probabilities, and then uses the current observation to update the state and posterior probability of each mode. Because each mode maintains a different state-transition hypothesis, changing driving patterns can be represented more flexibly than with a single model. The mode posterior probabilities indicate how well each mode explains the observations and are used as weights in the final state fusion. This property is suitable for representing trajectories exhibiting different path-speed patterns, such as straight driving, turning, deceleration, and avoidance, as mode-specific hypotheses. In this study, the Kalman filter and IMM state updates are not used for state estimation or trajectory post-processing during inference; they are used only for offline generation of path teacher labels, speed teacher labels, and mode posterior probabilities from expert driving trajectories.

## III. PROPOSED METHOD

### 1. Path-Speed Trajectory Output Representation

This study represents the future trajectory output of an end-to-end autonomous driving model as a path-speed trajectory composed of path and speed information. At time $t$, the expert driving trajectory $Y_t$ and the trajectory predicted by the learning model $\hat{Y}_t$ are defined as in (1).

$$Y_t = \{P_t, V_t\},\ \ \hat{Y}_t = \{\hat{P}_t, \hat{V}_t\} \tag{1}$$

In (1), $P_t$ and $\hat{P}_t$ are the expert and predicted path waypoint sequences, respectively, while $V_t$ and $\hat{V}_t$ are the corresponding speed waypoint sequences.

### 2. Offline IMM-Based Teacher-Signal Generation

Offline IMM-based teacher-signal generation produces path teacher labels, speed teacher labels, and mode posterior probabilities from expert driving trajectories. The expert trajectories are converted into the path-speed representation and then used in the IMM state update; the fused mode-specific states yield the path and speed teacher labels. The fixed-transition-probability IMM and SC-IMM use the same state representation and update structure, but SC-IMM adjusts the mode-transition probabilities and process noise on a frame-by-frame basis according to driving-context cues.

#### 2.1 Path-Speed State Representation

The future waypoint sequence is converted into a path-speed observation to summarize the path geometry and speed profile in a low-dimensional representation. In the ego-vehicle coordinate system, the future path is approximated by a quadratic curve with respect to the longitudinal coordinate $s$, and the speed profile is approximated by a linear model with respect to the prediction time $\tau$.

$$y(s) = c_0 + c_1 s + c_2 s^2,\ \ v(\tau) = v_{\mathrm{ref}} + a_{\mathrm{ref}}\tau \tag{2}$$

In (2), $c_0$, $c_1$, and $c_2$ are the path-shape coefficients, while $v_{\mathrm{ref}}$ and $a_{\mathrm{ref}}$ denote the reference speed and acceleration/deceleration tendency of the speed profile, respectively. The path and speed approximation results are then represented by the path-speed observation $z_t$ and IMM state $x_t$ in (3).

$$z_t = [c_0, c_1, c_2, v_{\mathrm{ref}}, a_{\mathrm{ref}}]^T,\ \ x_t = [c_0, c_1, c_2, v, a]^T \tag{3}$$

This five-dimensional path-speed representation expresses path geometry and speed changes jointly in a single state space. It therefore enables the subsequent IMM state update to account for both the path structure and speed-change characteristics.

#### 2.2 Definition of Driving Modes

For IMM-based teacher-signal generation, recurring patterns in expert driving trajectories are represented by five latent driving modes: Cruise, Turn, Deceleration, Avoidance, and Recovery. Cruise corresponds to stable straight driving or lane keeping, Turn to curved driving or intersection turning, and Deceleration to slowing or stopping. Avoidance represents evasive behavior caused by hazards or route changes, while Recovery represents returning to a stable path after a lateral deviation. Each mode expresses a distinct path-speed transition hypothesis, and the mode-specific states and posterior probabilities are updated jointly during the IMM state update.

#### 2.3 IMM State Update and Teacher-Signal Generation

The IMM state update recursively updates the mode-specific path-speed states and posterior probabilities using the path-speed observation. Each driving mode $j$ maintains a mode-specific state $x_t^j$ corresponding to a distinct path-speed transition hypothesis. The prior mode probabilities for the current frame are computed from the posterior mode probabilities of the previous frame and the transition-probability matrix.

$$\bar{\mu}_t^j = \sum_i \mu_{t-1}^i\, \Pi_t^{ij} \tag{4}$$

In (4), $\bar{\mu}_t^j$ is the prior probability of mode $j$ in the current frame, and $\Pi_t^{ij}$ is the probability of transitioning from mode $i$ to mode $j$. After the prior mode probabilities are computed, Kalman-filter prediction and update are performed for each mode. The agreement between the current path-speed observation and the predicted state of mode $j$ is expressed by the innovation likelihood and computed as in (5).

$$l_t^j = \mathcal{N}(v_t^j; 0, S_t^j) \tag{5}$$

$v_t^j$ and $S_t^j$ denote the innovation and innovation covariance of mode $j$, respectively. The prior mode probability represents transition feasibility, while the innovation likelihood represents the ability of each mode to explain the current observation. These two quantities are combined to compute the posterior mode probability in (6) [14,15].

$$\mu_t^j = \frac{\bar{\mu}_t^j l_t^j}{\sum_r \bar{\mu}_t^r\, l_t^r} \tag{6}$$

The final fused state is computed as the posterior-probability-weighted sum of the mode-specific states.

$$x_t^T = \sum_j \mu_t^j\, x_t^j \tag{7}$$

In (7), $x_t^T$ is the teacher state that summarizes the expert driving trajectory in the path-speed state space for the current frame. The path coefficients and speed states are converted into path and speed waypoint sequences and used as the path and speed teacher labels, respectively. The mode posterior probabilities in (6) are used as auxiliary mode-learning signals during training.

In the IMM state update used in this study, the state-transition and observation models are linear to simplify frame-to-frame changes in the path-speed state. The path-speed observation is defined to have the same dimension as the state vector, allowing the observation matrix to be set to the identity matrix. The initial covariance, observation-noise covariance, and base process noise for each mode are represented by diagonal matrices. Their elements are determined by considering the relative scales of the path coefficients and speed states, as well as the mode-specific path-speed transition characteristics.

### 3. Adjustment of Transition Probabilities and Process Noise in SC-IMM

A fixed-transition-probability IMM uses the same mode-transition probabilities and process noise in every frame. In contrast, SC-IMM retains the same path-speed state representation and IMM update structure while adjusting the mode-transition probabilities and process noise using driving-context cues from the current frame. Fig. 1 illustrates the offline teacher-signal generation architecture of SC-IMM. The posterior mode probabilities and mode-specific states from the previous frame are mixed through the interaction process, and the current path-speed observation is used to update the mode-specific Kalman filters. Simultaneously, scene cues are incorporated into the adjustment of the transition probabilities and process noise. The resulting mode-specific states and posterior probabilities are combined to generate the path teacher labels, speed teacher labels, and mode posterior probabilities.

As summarized in Table 1, the scene cues used by SC-IMM comprise route commands, path geometry, speed changes, hazards, and recovery cues. These cues summarize the driving context in the current frame and serve as conditioning variables for adjusting the mode-transition probabilities and process noise during the IMM state update.

SC-IMM begins with the base transition-probability matrix $\Pi_0$ and computes the frame-specific transition-probability matrix using the scene cues $s_t$ and the corresponding adjustment term $A(s_t)$.

$$\Pi_t = \text{softmax}_{\text{row}}(\log \Pi_0 + A(s_t)) \tag{8}$$

The row-wise softmax in (8) normalizes each row so that its transition probabilities sum to one. The scene cues $s_t$ include route commands, curvature, speed reduction, hazards, route changes, and recovery cues. Increased curvature and turn commands increase the likelihood of transitioning to the Turn mode, while speed reduction and signal-related hazards increase the likelihood of transitioning to the Deceleration mode. Vehicle or pedestrian hazards and route changes are associated with the Avoidance mode, whereas cues indicating a return to a stable path after a lateral deviation are associated with the Recovery mode.

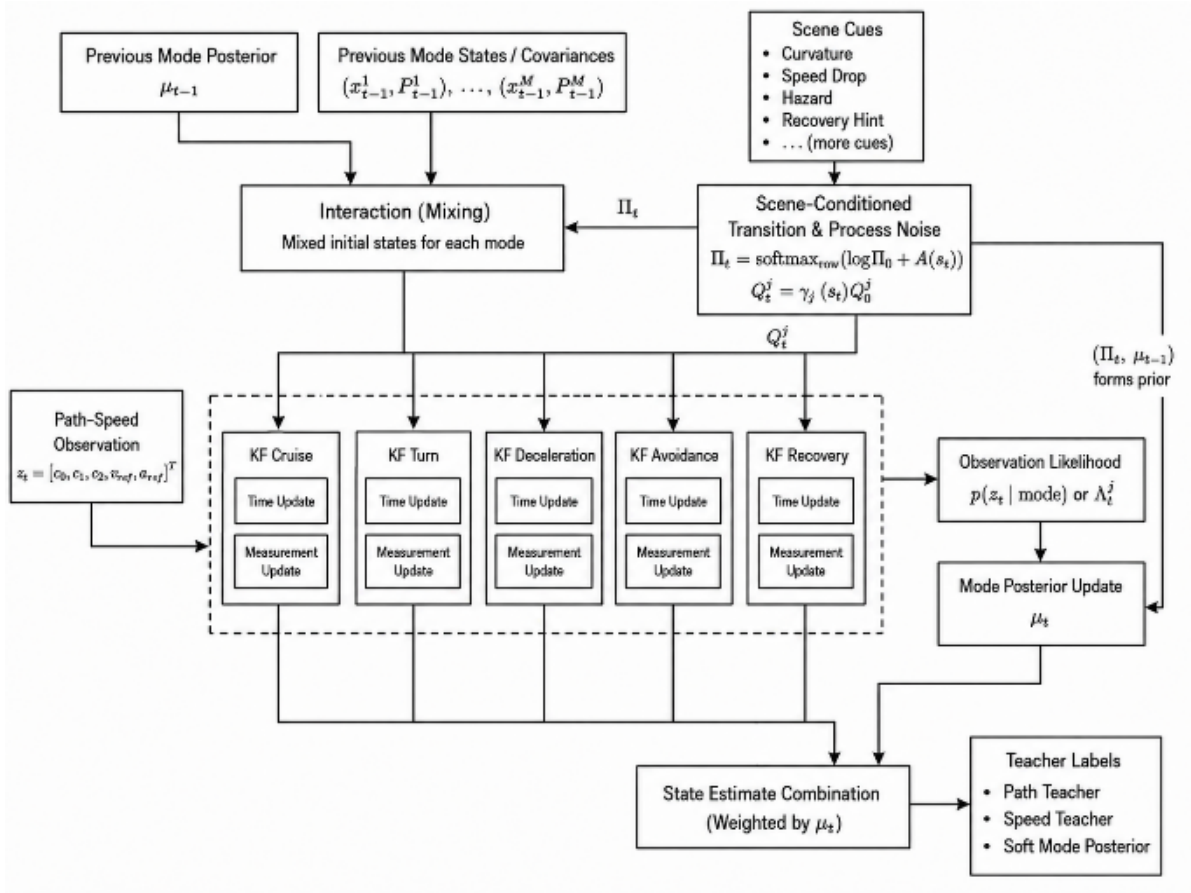

Fig. 1. SC-IMM-based offline teacher signal generation process.

Table 1. Scene cues used for SC-IMM based teacher signal generation.

| Cue Group | Variables | Role |
|---|---|---|
| Navigation cue | command, next command, junction | Route context |
| Path cue | curvature, curvature change, lateral offset, lateral change | Path geometry |
| Speed cue | ego speed, target speed, speed drop | Speed tendency |
| Hazard cue | vehicle, walker, traffic light, stop sign | Deceleration or avoidance cue |
| Recovery cue | changed route, recovery hint | Recovery behavior cue |

SC-IMM adjusts not only the transition probabilities but also the mode-specific process noise according to scene cues. When hazards, route changes, or abrupt speed reductions are present, it increases the prediction uncertainty of the corresponding mode, enabling the Kalman-filter prediction to respond more flexibly to abrupt changes in the driving situation.

$$Q_t^j = \gamma_j(s_t) Q_0^j \tag{9}$$

In (9), $Q_0^j$ is the base process noise of mode $j$, and $\gamma_j(s_t)$ is the uncertainty-adjustment coefficient determined by the scene cues. The final mode posterior probabilities of SC-IMM are computed by combining the prior mode probabilities obtained from the scene-conditioned transition probabilities with the mode-specific innovation likelihoods for the path-speed observation. The scene cues act as conditioning variables that adjust transition feasibility and prediction uncertainty, and the mode posterior is defined as a soft posterior that jointly reflects observations extracted from the expert trajectory and the driving-context cues.

### 4. Teacher-Signal-Based Learning Architecture

The learning model in this study is based on SimLingo-BASE [7], and the model trained only with the expert-trajectory loss is defined as the baseline. Fig. 2 shows the learning architecture with an auxiliary mode predictor. The model receives a front-view image, target point, and ego speed as inputs and outputs path and speed waypoints. The front-view image is converted into visual tokens, while the target point and ego speed are converted into conditioning tokens. The LLaMA backbone processes the visual tokens, conditioning tokens, and learnable driving queries; the path- and speed-query outputs are converted into path and speed waypoints through the waypoint predictors. While retaining the input-output format and waypoint-output structure of SimLingo-BASE, we add an auxiliary mode predictor that uses the average-pooled representations of the path- and speed-query outputs. The auxiliary mode predictor estimates logits for the five driving modes, and the posterior probabilities generated by SC-IMM are used in the auxiliary learning loss.

As shown in (10), the training loss is defined as a weighted sum of the original trajectory loss, the SC-IMM-based teacher-label loss, and the auxiliary mode-prediction loss.

$$L = \lambda_{\text{origin}} L_{\text{origin}} + \lambda_{\text{scimm}} L_{\text{scimm}} + \lambda_{\text{mode}} L_{\text{mode}} \tag{10}$$

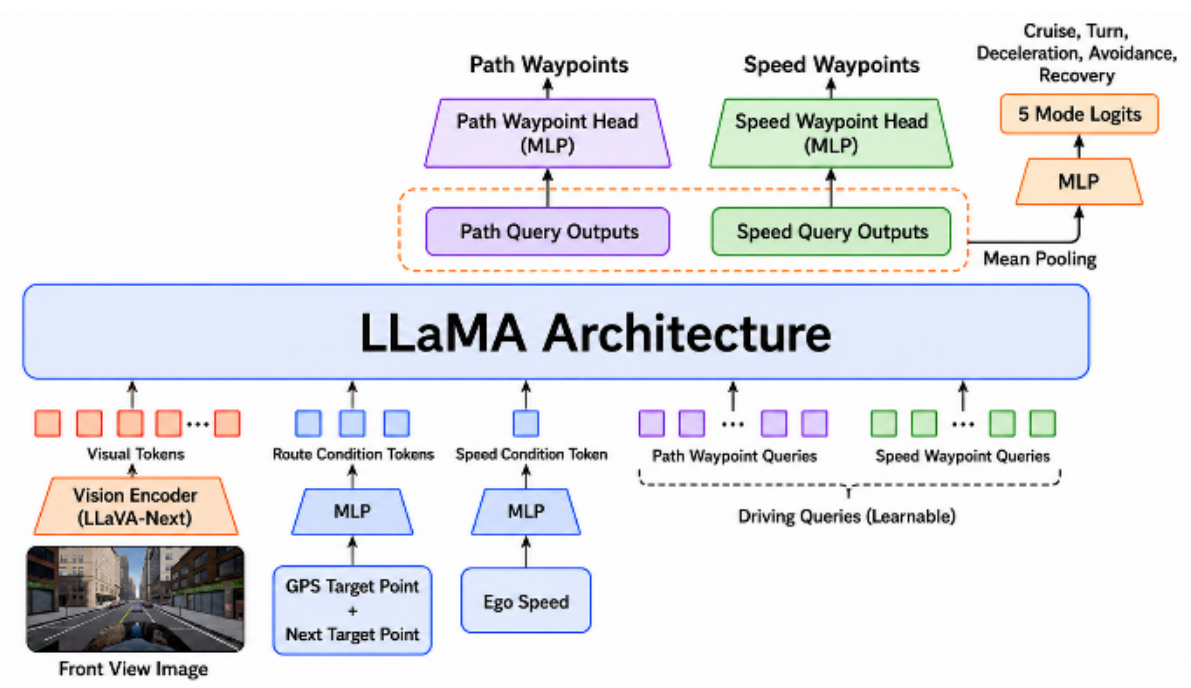

Fig. 2. Student model architecture with an auxiliary mode predictor.

In (10), $L_{\text{origin}}$ is the original loss that uses the expert driving trajectory as the learning target, while $L_{\text{scimm}}$ and $L_{\text{mode}}$ are the SC-IMM-based teacher-label loss and auxiliary mode-prediction loss, respectively. $L_{\text{scimm}}$ is an auxiliary loss that encourages the path and speed outputs of the learning model to follow the teacher labels generated by SC-IMM, and is defined as in (11).

$$L_{\text{scimm}} = \| \hat{P}_t - P_t^T \|_2^2 + \| \phi(\widehat{W}_t) - V_t^T \|_2^2 \tag{11}$$

$P_t^T$ and $V_t^T$ are the path and speed teacher labels generated by SC-IMM, respectively. Because the speed teacher label is represented as a speed profile, the waypoint output $\widehat{W}_t$ of the learning model is converted into the speed profile in (12) through the transformation function $\phi(\cdot)$ before comparison.

$$\phi(\widehat{W}_t)^k = \frac{\| \widehat{w}_t^k - \widehat{w}_t^{k-1} \|_2}{\Delta t} \tag{12}$$

Equation (12) divides the displacement between adjacent waypoints by the time interval $\Delta t$ to compute the $k$-th speed value. The auxiliary mode-prediction loss is defined as the Kullback-Leibler (KL) divergence between the mode posterior probabilities generated by SC-IMM and the mode probabilities predicted by the learning model.

$$L_{\text{mode}} = D_{\text{KL}}(\mu_t^T \parallel q_t) \tag{13}$$

In (13), $\mu_t^T$ is the teacher-mode posterior that jointly reflects the path-speed observation and driving-context cues, and $q_t$ is the mode probability predicted by the auxiliary mode predictor. By using this posterior as the learning target, the auxiliary mode predictor encourages the shared driving representation to learn the relationship between path-speed changes and driving context. Its output is used only to compute the loss during training and is not used for waypoint prediction or control-command computation during inference.

## IV. EXPERIMENTS AND ANALYSIS

### 1. Evaluation Environment and Comparison Methods

Experiments were conducted in the CARLA simulator [19] to evaluate the closed-loop driving performance of the proposed method. The training data consisted of driving sequences collected in the CARLA environment using PDM-Lite [20], a rule-based agent. For a controlled comparison of driving performance and stability, we independently constructed 100 short routes in Town12, each approximately 300 m long. Fig. 3 shows the Town12 short-route configuration used for the evaluation.

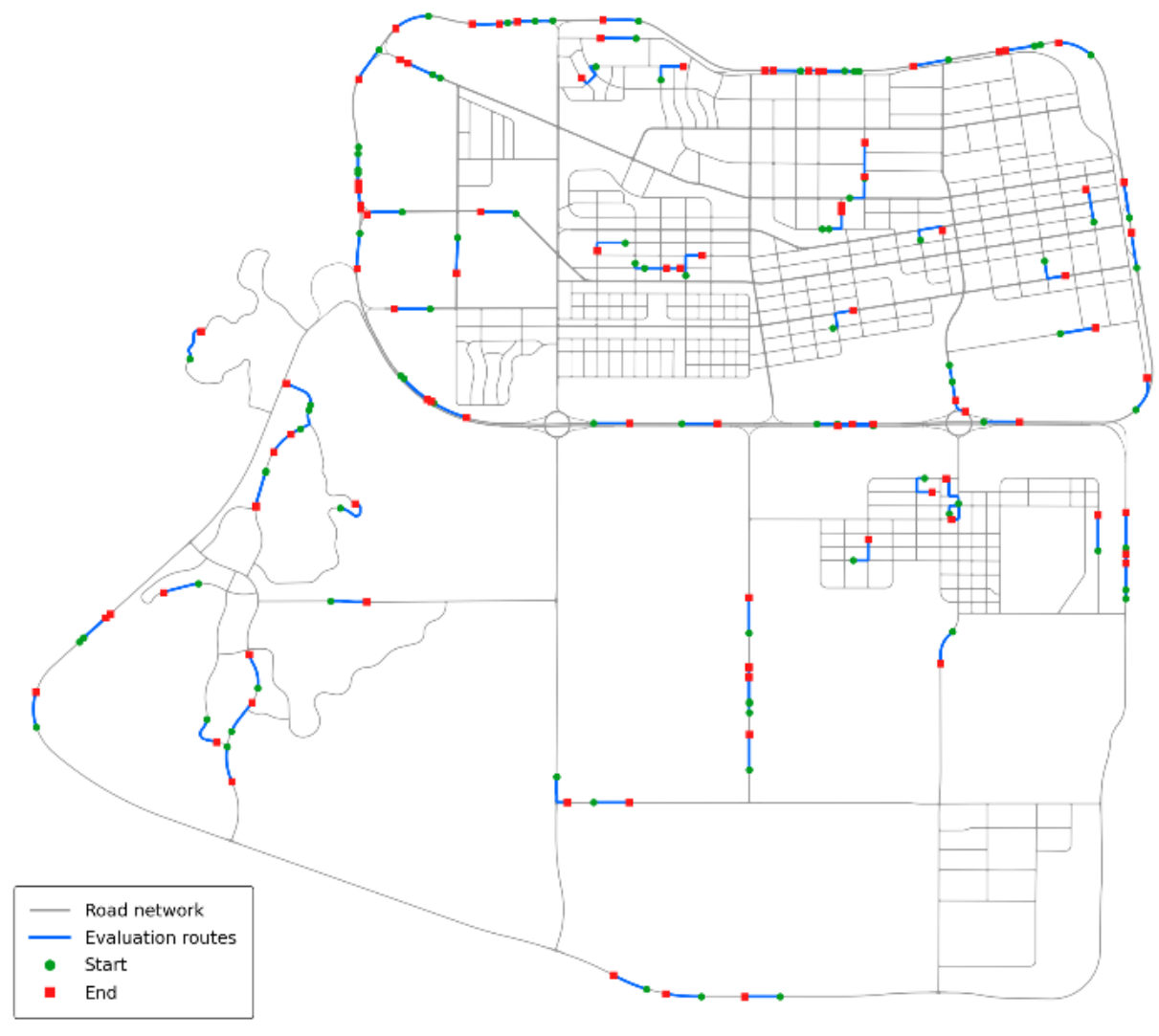

Fig. 3. Town12 short-route evaluation configuration.

The evaluated models comprised a baseline trained only with the expert-trajectory loss, a model that adds fixed-transition-probability IMM-based teacher signals to the baseline loss, and a model that adds SC-IMM-based teacher signals to the baseline loss. All models used the same evaluation routes, input information, and waypoint controller. No online IMM state update, trajectory post-processing, or mode-conditioned control was applied during evaluation.

Driving performance was evaluated according to the CARLA Leaderboard 2.0 protocol [21] using Driving Score (DS), Route Completion (RC), and Infraction Score (IS). RC is the proportion of the route completed by the vehicle, while IS corresponds to the infraction penalty of the CARLA Leaderboard. DS is a composite score that reflects both RC and IS, evaluating not only route completion but also stable driving without infractions. For the complete set of $N$ routes, DS, RC, and IS are defined as in (14).

$$\begin{aligned} DS &= \frac{1}{N}\sum_{i=1}^{N} R_i P_i, \\ RC &= \frac{1}{N}\sum_{i=1}^{N} R_i, \\ IS &= \frac{1}{N}\sum_{i=1}^{N} P_i. \end{aligned} \tag{14}$$

In (14), $R_i$ and $P_i$ are the route completion and infraction penalty, respectively, for the $i$-th route. Collision frequency was evaluated as Collision/km, the number of collisions per distance traveled. Temporal changes in the control commands were evaluated using steering jerk and longitudinal jerk, while temporal changes in trajectory outputs were evaluated using curvature variation and speed variation. Lower values of all Collision/km, jerk, and variation metrics indicate lower collision frequency or smaller temporal changes. Performance improvements were computed as relative changes from the baseline, accounting for whether a higher or lower value indicates better performance.

Table 2. Mode-wise path-speed variation metrics of raw expert trajectories and teacher labels.

| Mode | Method | Speed Δ ↓ | Speed Jerk ↓ | Lat Δ ↓ | Heading Δ ↓ | Lat Jerk ↓ |
|---|---|---|---|---|---|---|
| Cruise | Original Raw Expert | 0.308 | 9.49 | 0.00776 | 0.00101 | 0.260 |
| Cruise | Fixed-IMM | **0.142** | 3.18 | **0.00356** | 0.000505 | **0.102** |
| Cruise | SC-IMM | **0.142** | **3.05** | 0.00380 | **0.000497** | **0.102** |
| Turn | Original Raw Expert | 0.267 | 9.94 | 0.0179 | 0.00356 | 0.432 |
| Turn | Fixed-IMM | 0.127 | 2.78 | 0.0104 | **0.00215** | **0.216** |
| Turn | SC-IMM | **0.120** | **1.74** | **0.00992** | 0.00217 | **0.216** |
| Deceleration | Original Raw Expert | 0.864 | 18.00 | 0.0147 | 0.00235 | 0.429 |
| Deceleration | Fixed-IMM | **0.491** | 7.01 | 0.00821 | 0.00137 | 0.197 |
| Deceleration | SC-IMM | 0.510 | **6.91** | **0.00757** | **0.00128** | **0.188** |
| Avoidance/Recovery | Original Raw Expert | 0.583 | 16.36 | 0.0560 | 0.00839 | 1.769 |
| Avoidance/Recovery | Fixed-IMM | 0.309 | 5.52 | **0.0269** | **0.00417** | 0.738 |
| Avoidance/Recovery | SC-IMM | **0.294** | **4.95** | 0.0298 | 0.00426 | **0.670** |

## 2. Analysis of SC-IMM-Based Teacher Signals

### 2.1 Mode-Wise Characteristics of Path-Speed Teacher Labels

The characteristics of the teacher labels were analyzed in terms of mode-wise speed change, lateral change, heading change, and jerk. Table 2 compares the original expert trajectories with the teacher labels from the two IMM methods in Cruise, Turn, Deceleration, and Avoidance/Recovery segments. Avoidance and Recovery were combined because they frequently occur consecutively during lateral adjustment after a hazard or route change. A lower value for each metric indicates a smaller path or speed change between consecutive time steps.

Compared with the original expert trajectories, SC-IMM generally reduced the speed- and lateral-change metrics in every driving mode. This indicates that the SC-IMM-based teacher labels attenuated abrupt path and speed changes in the original expert trajectories into more stable forms. Compared with the fixed-transition-probability IMM, SC-IMM also produced lower Speed Jerk in every mode. Abrupt speed changes were particularly reduced in the Turn and Avoidance/Recovery segments, while Lat Δ, Heading Δ, and Lat Jerk all decreased in the Deceleration segment. This result can be attributed to the ability of SC-IMM to apply driving-context cues according to the characteristics of segments in which path and speed changes occur jointly, such as deceleration, turning, and avoidance. It also suggests that SC-IMM preserves necessary path changes while reducing unnecessary instantaneous fluctuations. In some segments, however, SC-IMM tended to preserve path changes required for turning, avoidance, and recovery rather than minimizing Lat Δ or Heading Δ. These results indicate that the SC-IMM-based teacher labels are not merely smoothed trajectories, but can serve as auxiliary learning signals that reduce speed and lateral jerk while reflecting driving-context cues.

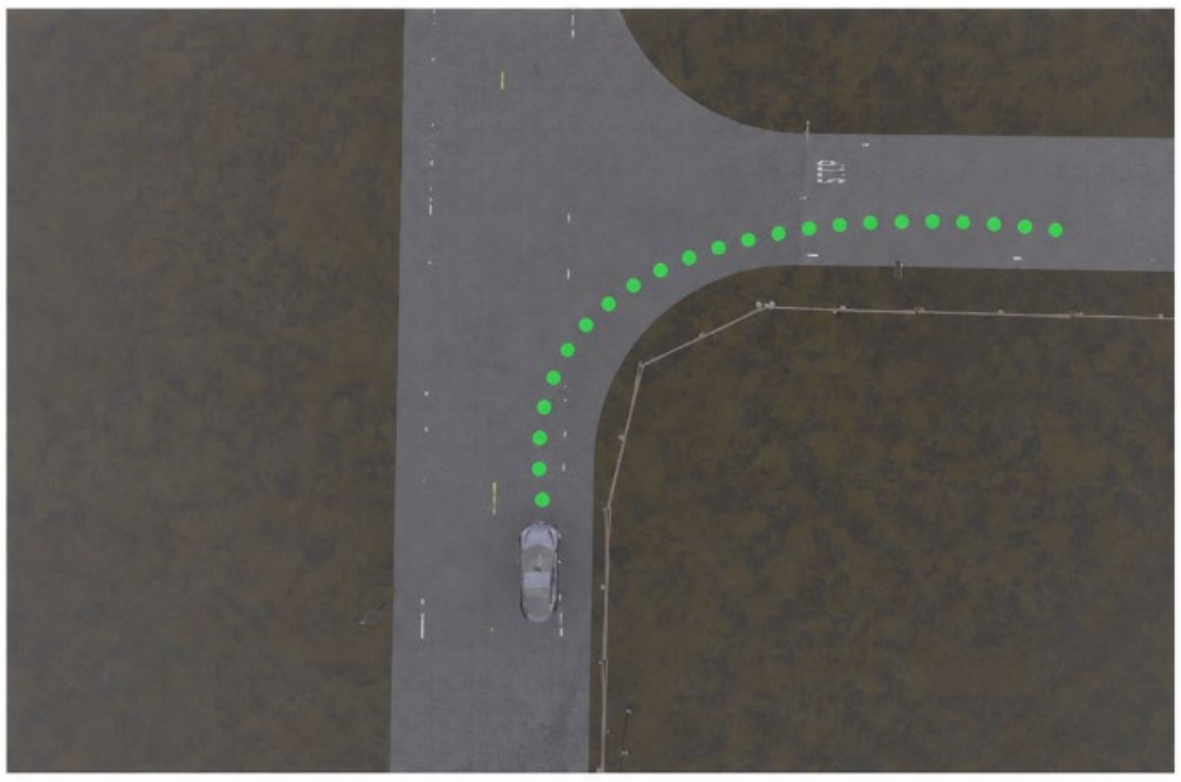

(a) BEV visualization of the right-turn driving sequence

(b) Temporal variation of SC-IMM mode posterior

Fig. 4. SC-IMM mode posterior in a right-turn sequence.

### 2.2 Analysis of Mode Posterior Characteristics

The mode-posterior characteristics were analyzed by jointly examining scene changes and frame-wise mode-probability changes in a right-turn driving sequence. Fig. 4(a) visualizes the right-turn sequence from a bird's-eye-view perspective, while Fig. 4(b) shows the posterior probabilities of the Cruise, Turn, Deceleration, Avoidance, and Recovery modes computed through the SC-IMM state update for the same sequence. The mode with the highest posterior probability in each frame is the dominant driving mode at that time, while the entire probability distribution is used as soft supervision in the mode-prediction loss.

The sequence in Fig. 4(a) shows the vehicle decelerating, entering a right-turn segment, and then continuing along the route. Correspondingly, Fig. 4(b) shows a relatively high Deceleration posterior in the initial segment, followed by an increase in the Turn posterior during the right turn. After the turn, the Cruise posterior remains relatively high. These results show that the SC-IMM mode posterior varies continuously with scene and path-speed changes throughout the driving sequence and serves as a teacher signal expressed as a context-dependent probability distribution rather than a single hard mode label.

## 3. Closed-Loop Driving Performance Analysis

### 3.1 Quantitative Analysis of Driving Performance and Temporal Stability

Closed-loop driving performance was compared among the baseline, fixed-transition-probability IMM, and SC-IMM methods. The baseline was trained only with the expert-trajectory loss, whereas the fixed-transition-probability IMM and SC-IMM models add the corresponding teacher signals to the baseline training loss. All three models used the same evaluation routes, input information, and waypoint controller, and neither online IMM state updates nor mode-conditioned control was applied during evaluation.

Table 3 presents the closed-loop driving performance on the Town12 evaluation routes. Because all three models achieved an RC of 100.00, the performance differences are mainly reflected in the number of infractions and collisions. Both teacher-signal models improved closed-loop driving performance relative to the baseline. The fixed-transition-probability IMM achieved higher DS and IS values than the baseline and substantially reduced Collision/km. SC-IMM achieved the highest DS and IS and the lowest Collision/km among the three models. In particular, SC-IMM improved DS by 28.0% and reduced Collision/km by 62.3% relative to the baseline. This suggests that the SC-IMM-based teacher signals provide context-dependent path-speed transition characteristics during training and thereby improve closed-loop driving performance without modifying the inference architecture.

Table 4 presents the temporal-stability metrics for the control and trajectory outputs measured on the same evaluation routes. Steering jerk and longitudinal jerk describe temporal changes in the control commands, while curvature variation and speed variation describe changes in the curvature and speed of trajectories output over consecutive frames. SC-IMM also generally improved the temporal-stability metrics. Steering jerk, longitudinal jerk, and curvature variation were all lower than those of both the baseline and fixed-transition-probability IMM. This indicates that training with SC-IMM-based teacher signals helps attenuate changes in predicted trajectory direction and curvature across consecutive frames, as well as the resulting changes in control input. Although the fixed-transition-probability IMM achieved the lowest speed variation, SC-IMM also substantially reduced it relative to the baseline.

These results indicate that SC-IMM does not merely minimize smoothness metrics; instead, it preserves path changes required by the driving context while suppressing unnecessary frame-to-frame fluctuations.

Table 3. Closed-loop driving performance on Town12 evaluation routes.

| Method | DS ↑ | RC ↑ | IS ↑ | Collision /km ↓ |
|---|---|---|---|---|
| Baseline | 58.70 | **100.00** | 0.587 | 5.055 |
| Fixed-IMM | 71.82 | **100.00** | 0.718 | 2.042 |
| SC-IMM | **75.11** | **100.00** | **0.751** | **1.908** |

Table 4. Control and trajectory-output stability on Town12 evaluation routes.

| Method | Steering. jerk ↓ | Long. jerk ↓ | Curv. var. ↓ | Speed var. ↓ |
|---|---|---|---|---|
| Baseline | 19.11 | 44.01 | 0.0430 | 4.662 |
| Fixed-IMM | 17.84 | 40.80 | 0.0337 | **3.200** |
| SC-IMM | **16.13** | **38.96** | **0.0289** | 3.317 |

In particular, the reductions in steering jerk and curvature variation can be interpreted as factors that reduce abrupt changes in the control input generated by the waypoint controller and make vehicle behavior more consistent. Thus, training with SC-IMM-based teacher signals can improve both closed-loop driving performance and trajectory-output stability without online filtering or controller modifications during inference.

### 3.2 Qualitative Analysis of Waypoint-Output Stability

Fig. 5 overlays the future waypoint sequences output by the baseline and SC-IMM models in an ego-centered bird's-eye-view coordinate system over the same closed-loop driving segment. In each panel, the light-colored lines represent waypoint sequences predicted in consecutive frames, while the dark line represents the waypoint sequence output in the final frame of the segment.

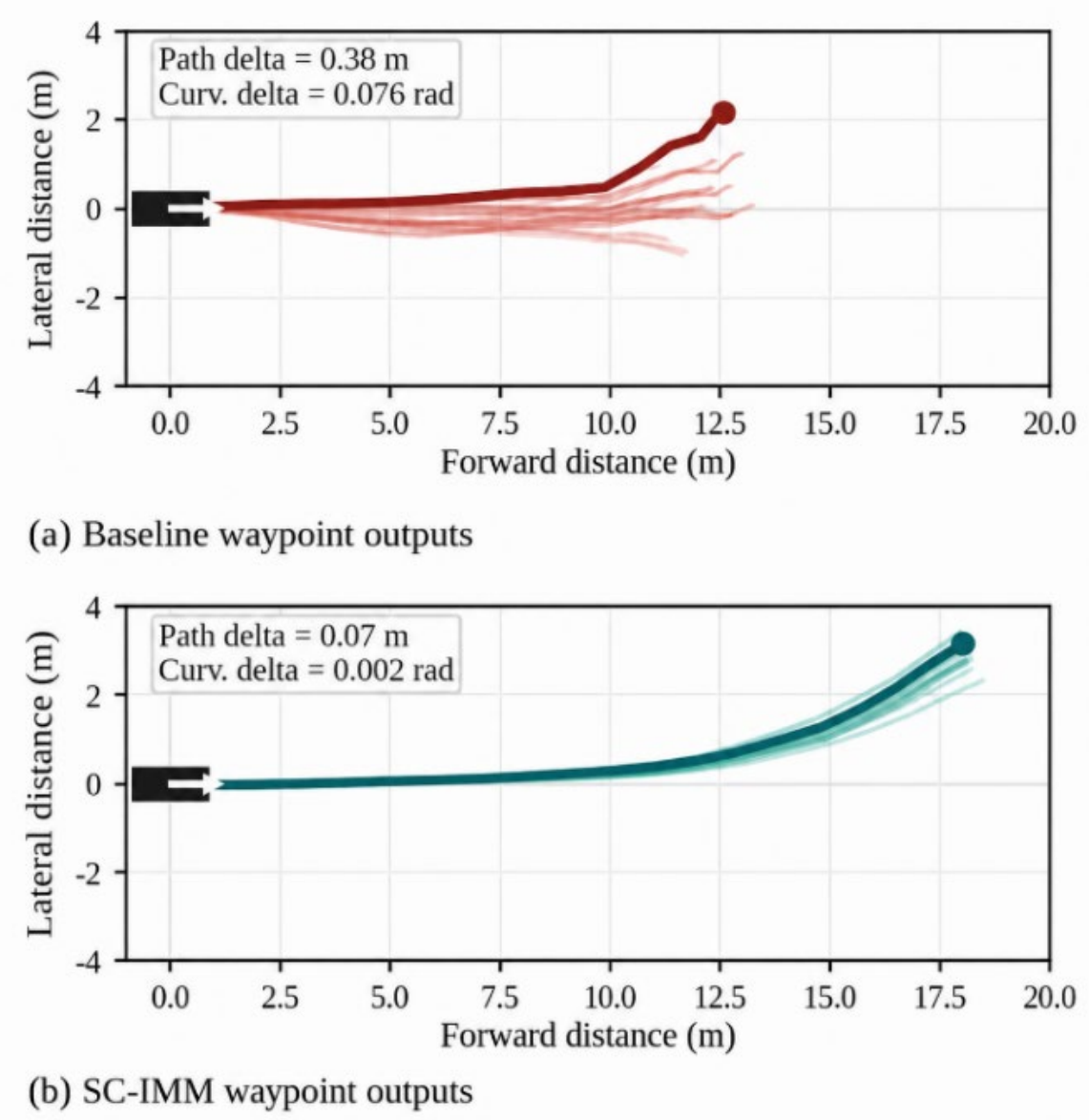

Fig. 5. Qualitative comparison of waypoint-output stability.

The baseline in Fig. 5(a) exhibits relatively large changes in the lateral position and curvature of the waypoints between adjacent frames, and its waypoint sequences are widely dispersed across consecutive frames. In contrast, the consecutive waypoint sequences of the SC-IMM model in Fig. 5(b) overlap more closely, while the path geometry and curvature remain comparatively stable. This demonstrates that training with SC-IMM-based teacher signals reduces frame-to-frame variation in future waypoint outputs during closed-loop driving.

## V. CONCLUSION

This paper proposed an SC-IMM-based offline teacher-signal generation and learning method to improve the stability of trajectory outputs from end-to-end autonomous driving models. The proposed method converts expert driving trajectories into path-speed states and generates path teacher labels, speed teacher labels, and mode posterior probabilities through IMM state updates. The path and speed teacher labels serve as learning targets for trajectory-output training, while the mode posterior probabilities provide soft supervision for auxiliary mode learning. The generated teacher signals are used only during training; no online filtering, trajectory post-processing, or mode-conditioned control is performed during inference.

Closed-loop evaluation on a separately constructed set of short routes in CARLA Town12 showed that the SC-IMM-trained model generally improved both driving performance and temporal stability compared with the baseline and fixed-transition-probability IMM. Quantitative evaluation confirmed improvements in driving score, collision frequency, and the temporal-change metrics of the control and trajectory outputs. The qualitative analysis also showed that waypoint outputs remained more consistent across consecutive frames. These results demonstrate that path-speed teacher labels and mode posterior probabilities conditioned on driving-context cues can serve as auxiliary learning signals for stabilizing the trajectory outputs of end-to-end autonomous driving models.

Future work will examine the generalization of SC-IMM-based teacher signals across a broader range of road geometries and traffic situations. We also plan to refine the scene-cue-based adjustment of transition probabilities and process noise, and to further analyze its effects on long-term driving stability in diverse simulation environments and real-vehicle settings.

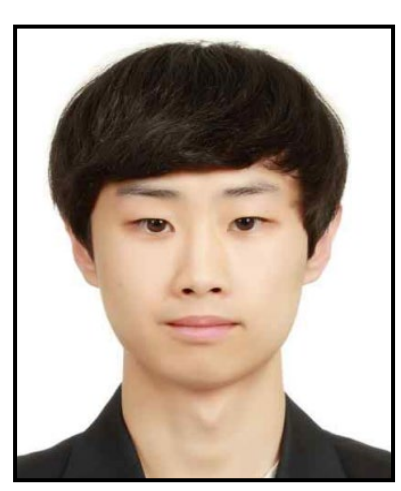

**Siewoo Kim**
Siewoo Kim received the B.S. degree in computer engineering from Hongik University in 2025. He is currently pursuing the M.S. degree at The CCS Graduate School of Mobility, Korea Advanced Institute of Science and Technology. His research interests include end-to-end autonomous driving, robotics, and computer vision.

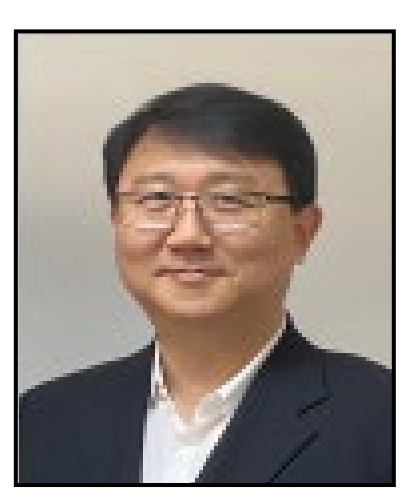

**Seung-Hyun Kong**
Seung-Hyun Kong received the B.S. degree in electronic engineering from Sogang University in 1992, the M.S. degree in electrical engineering from New York University in 1994, and the Ph.D. degree in aeronautics and astronautics from Stanford University in 2006. From 1997 to 2004, he was with the Telecommunications R&D Center at Samsung Electronics, and from 2006 to 2010, he was a Senior Researcher at Qualcomm R&D in the United States. Since 2010, he has been an Associate Professor at The CCS Graduate School of Mobility, Korea Advanced Institute of Science and Technology (KAIST). His research interests include autonomous driving, vehicle positioning, GNSS and navigation technologies, deep learning, and sensor fusion.